\documentclass[10pt,conference]{IEEEtran}
\usepackage{cite}
\usepackage[dvips]{color}
\usepackage{tabu}
\usepackage{comment}
\usepackage{todonotes}
\usepackage{epsf}
\usepackage{epsfig}
\usepackage{times}
\usepackage{epsfig}
\usepackage{graphicx}

\usepackage{mathtools}
\usepackage{mathrsfs}
\usepackage{amssymb,amsmath}
\usepackage{amsmath}
\usepackage{amsfonts}
\usepackage{enumitem} 

\usepackage{algorithmicx}  
\usepackage[algo2e]{algorithm2e} 
\usepackage{cleveref}
\usepackage{dsfont}
\usepackage{lettrine} 
\usepackage{epsfig,algorithm,algpseudocode,amsthm,url}
\usepackage{caption}
\usepackage{subcaption}
\usepackage[justification=raggedright]{caption}
\usepackage{subcaption}
\usepackage{bbm}
\usepackage[utf8]{inputenc}
\usepackage{verbatim}
\usepackage{setspace}	
\usepackage[T1]{fontenc}
\usepackage{textcomp}
\usepackage{tabularx}
\usepackage{xpatch}
\DeclareMathOperator*{\argmax}{argmax}

\usepackage{stackengine}
\usepackage[nocomma]{optidef}
\usepackage{commath}

\usepackage[left=1.62cm,right=1.62cm,top=1.85cm,bottom=2.6cm]{geometry}

\IEEEoverridecommandlockouts 

  \renewcommand{\baselinestretch}{0.91}

\begin{document}
\title{\Huge Prompt-Tuned LLM-Augmented DRL for Dynamic O-RAN Network Slicing 
\thanks{This material is based upon work supported by the National Science Foundation under Grant Numbers  CNS-2202972, CNS- 2318726, and CNS-2232048.} 
}\vspace{-0.3cm}

\author{
	\IEEEauthorblockN{
	Fatemeh Lotfi, Hossein Rajoli, Fatemeh Afghah}

	\IEEEauthorblockA{Holcombe Department of Electrical and Computer Engineering, Clemson University, Clemson, SC, USA \\
Emails: flotfi@clemson.edu, hrajoli@clemson.edu,  fafghah@clemson.edu} }\vspace{-0.2cm}

\maketitle\vspace{-0.5cm}
\begin{abstract} 
Modern wireless networks must adapt to dynamic conditions while efficiently managing diverse service demands. Traditional deep reinforcement learning (DRL) struggles in these environments, as scattered and evolving feedback makes optimal decision-making challenging. Large Language Models (LLMs) offer a solution by structuring unorganized network feedback into meaningful latent representations, helping RL agents recognize patterns more effectively. For example, in O-RAN slicing, concepts like SNR, power levels and throughput are semantically related, and LLMs can naturally cluster them, providing a more interpretable state representation. 
To leverage this capability, we introduce a contextualization-based adaptation method that integrates learnable prompts into an LLM-augmented DRL framework. Instead of relying on full model fine-tuning, we refine state representations through task-specific prompts that dynamically adjust to network conditions. Utilizing ORANSight, an LLM trained on O-RAN knowledge, we develop Prompt-Augmented Multi agent RL (PA-MRL) framework. Learnable prompts optimize both semantic clustering and RL objectives, allowing RL agents to achieve higher rewards in fewer iterations and adapt more efficiently. 
By incorporating prompt-augmented learning, our approach enables faster, more scalable, and adaptive resource allocation in O-RAN slicing. Experimental results show that it accelerates convergence and outperforms other baselines.
\end{abstract}\vspace{-0.2cm}

\section{Introduction}\label{intro}

Advanced wireless networks must be highly dynamic and capable of managing heterogeneous service demands efficiently. Network slicing technique is a key feature of these networks that enable them to support heterogeneous services like eMBB (enhanced Mobile Broadband), URLLC (Ultra-Reliable Low-Latency Communications), and mMTC (massive Machine-Type Communications). Recognizing its importance, the open radio access network (O-RAN) ALLIANCE has identified network slicing as a critical technology for building flexible and intelligent radio access networks~\cite{oran_wg3_2024}.

The modular nature of the O-RAN architecture through its separation of components like the RAN Intelligent Controller (RIC), Centralized Units (CUs), and Distributed Units (DUs), lays the foundation for scalable and distributed network control~\cite{3gppRe18,polese2022understanding}. 
This disaggregation makes it well-suited for deploying multi-agent reinforcement learning (MARL), where multiple deep reinforcement learning (DRL) agents operate across the CUs and DUs~\cite{lotfi2024metareinforcementlearningapproach,naderializadeh2021resource}. At the DU level, agents manage localized tasks such as resource block (RB) allocation and user scheduling, while inter-agent coordination enables system-wide optimization. 
Despite these advances, traditional DRL approaches often struggle to adapt quickly in dynamic and rapidly changing network conditions. They typically require numerous training iterations to establish reliable state-action mappings. Moreover, the scattered and noisy nature of real-time feedback in wireless environments complicates the agent’s ability to extract structured representations and make optimal decisions. 

Recent studies have attempted to augment deep learning approaches by incorporating Large Language Models (LLMs) into wireless systems, enhancing deep learning-based decision-making~\cite{merouane2024large,bariah2023understanding,chen2024communication,lotfi2025llm,melike2024llm,zhang2023controlling,wang2024llm}. 
Transfer learning, particularly with LLMs, plays a crucial role by enabling generalization across tasks through pretraining and fine-tuning. However, conventional fine-tuning demands substantial labeled data and computational resources and challenges that are particularly limiting in dynamic and resource constrained environments like O-RAN slicing. To mitigate this, more efficient adaptation strategies such as LoRA~\cite{hu2022lora} and adapter-based methods~\cite{houlsby2019parameter} have been proposed. These methods introduce lightweight trainable components into frozen models, allowing task adaptation with significantly fewer parameters. 
To further enhance efficiency and adaptability in O-RAN, this work adopts a lightweight and scalable prompt-learning LLM adaptation strategy that uses learnable prompts, small trainable embeddings that guide the LLM without modifying its core parameters. In contrast to traditional fine-tuning or adapter-based approaches, this approach offers a lightweight and scalable solution that is well-suited for real-time, decentralized wireless environments scenarios like O-RAN, where agents must adapt on the fly with limited feedback. 
We employ ORANSight, a specialized LLM pre-trained on comprehensive O-RAN knowledge~\cite{gajjar2025oransight,gajjar2024oran13k}, to contextualize observations into meaningful latent representations and transform scattered network feedback into semantically meaningful latent representations. These representations are used by decentralized DRL agents to make more adaptive and generalizable slicing and scheduling decisions. 

This work moves beyond traditional fine-tuning by proposing a novel contextualization-based framework that enhances decision-making in O-RAN slicing by coupling LLM-driven representations with MARL. Unlike traditional fine-tuning approaches that modify the entire language model, we introduce learnable prompts as compact and trainable tokens that guide the pre-trained LLM in generating semantically meaningful representations of the environment. These prompts are updated with the DRL agent's policy, ensuring that state representations align with the agent's learning objectives. This design allows the model to flexibly interpret complex network dynamics, such as varying QoS levels or RF conditions, without incurring the high costs of full model retraining. Additionally, by maintaining the pretrained LLM's core parameters, our method preserves its generalization capabilities while adapting efficiently to task-specific goals. Integrated into a MARL framework, this setup enables decentralized agents to develop faster, more robust policies for joint slicing and resource scheduling in O-RAN. As a result, our approach delivers improved adaptability, faster convergence, and lower computational overhead, making it well-suited for real-time and scalable wireless network management. 
The main contributions of this paper are summarized as follows: 
\begin{itemize}
    \item We propose a novel Prompt-Augmented Multi-Agent Reinforcement Learning (PA-MRL) framework tailored for O-RAN slicing, which integrates both learnable and informal prompts to enhance state representation and decision-making.
    \item We combine LLM-generated textual embeddings with raw network metrics using pre-trained adapter networks, forming a unified and semantically meaningful representation that improves agent awareness and convergence.
    \item By embedding prompt learning into the RL loop, our method enables faster learning and better generalization in highly dynamic, decentralized O-RAN environments, outperforming traditional fine-tuning approaches in efficiency and adaptability.
    \item We implement our approach using Soft Actor-Critic (SAC) in a MARL setup distributed across DUs, enabling fine-grained, scalable resource allocation while preserving stability in continuous action spaces.
\end{itemize}
 \emph{To the best of our knowledge, this is the first work to adaptively integrate context-aware LLM-based state representation into a MARL framework for joint O-RAN slicing and scheduling}.  
\begin{figure}[t!]
  \centering
    \includegraphics[width=0.75\columnwidth]{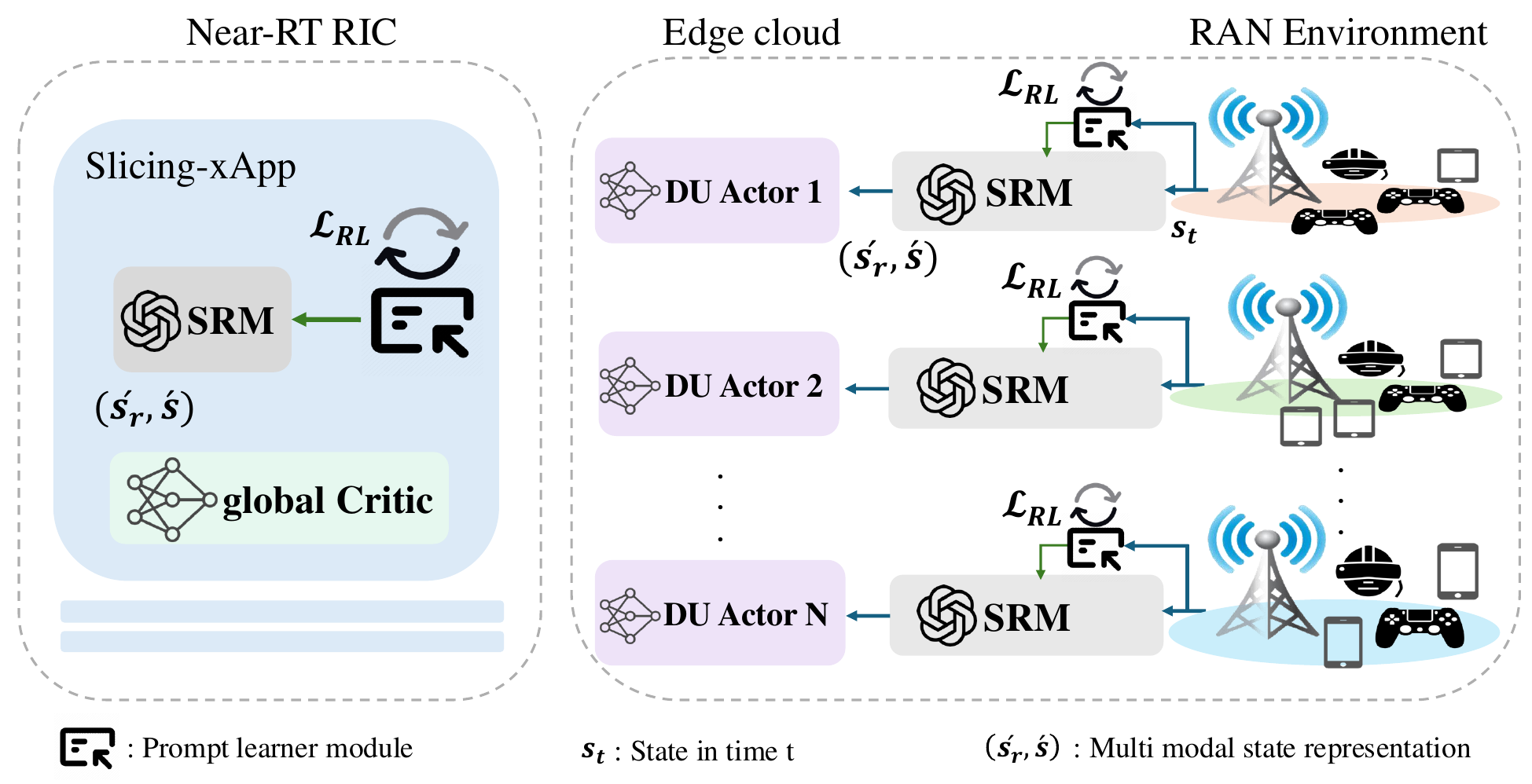}
  \caption{Prompt Enhancement LLM Augmented MARL System Framework. 
  }\vspace{-0.2cm}
    \label{sys_graph}
\end{figure}


\section{Related Works}\label{related_works}

LLMs have recently emerged as powerful tools in wireless communication research. Their capacity to handle and reason over diverse data types, such as textual inputs and RF signal measurements, has led to their growing application in wireless communication areas, including intelligent network control, anomaly detection, and analysis of 3GPP standards~\cite{lotfi2025llm,melike2024llm,gajjar2025oransight}.\vspace{-0.1cm}

\subsection{Transfer Learning Foundations}
Transfer learning enables deep learning models to apply pre-existing knowledge to new tasks, typically through pretraining on large-scale corpora followed by task-specific fine-tuning. Although effective, fine-tuning is resource-intensive and relies on labeled data, which limits its practicality in dynamic, low-resource settings like O-RAN slicing. To address this, recent studies propose parameter-efficient alternatives such as LoRA (Low-Rank Adaptation)~\cite{hu2022lora} and adapter modules~\cite{houlsby2019parameter}, which insert lightweight, trainable components into frozen models. Another alternative is prompt learning~\cite{lester2021power,li2021prefix}, where learnable prompts steer model outputs without modifying core weights, making it ideal for few-shot or zero-shot settings.\vspace{-0.1cm}

\subsection{LLM-Augmented RL in Wireless Network Settings}
Recent efforts have begun integrating LLMs with RL to support dynamic decision-making in wireless systems. For example, authors in ~\cite{chen2024communication} use LLMs to enhance state representation and feedback design in offline learning, though their approach lacks real-time adaptability. Also, authors in~\cite{melike2024llm} combine hierarchical RL with LLMs to translate operator intents into network actions, but the system remains reactive mainly and constrained by static intent structures. Other works~\cite{zhang2023controlling,wang2024llm} apply LLMs in multi-agent RL, using centralized critics and token-based feedback to improve coordination. 
However, both depend heavily on centralized architectures and static representations, which are less suitable for O-RAN’s decentralized, real-time nature.

Building on recent insights, our work proposes a prompt-augmented DRL framework tailored for dynamic O-RAN slicing. While prior studies have shown the potential of combining LLMs and RL, they often fall short in real-time adaptability, rely on centralized architectures, and overlook the value of multimodal data. To address these limitations, we integrate contextual prompts into the DRL loop, allowing decentralized agents to make more adaptive slicing decisions without the need for full model fine-tuning. This lightweight, flexible approach enhances generalization and responsiveness, making it well-suited for fast-changing RAN environments.\vspace{-0.2cm}



\section{System Model and Problem Formulation}\label{sys_model}

\subsection{Overall Architecture}
The O-RAN network architecture under consideration consists of three distinct slices, each designed to fulfill specific Quality of Service (QoS) requirements: eMBB for high-throughput applications, mMTC for massive connectivity, and URLLC for delay-sensitive services. 
At its core, the RIC orchestrates network resources by coordinating CUs, DUs, and RUs. Resource allocation operates at two levels; Inter-Slice Management, ensuring fair distribution across slices based on QoS, and Intra-Slice Management, allocating resources among UEs within each slice per scheduling policies. The MAC layer maps these allocations to UEs, aligning with network optimization goals. The overall architecture is depicted in Fig. \ref{sys_graph}.
\begin{figure}[t!]
  \centering
  \includegraphics[width=0.53\columnwidth]{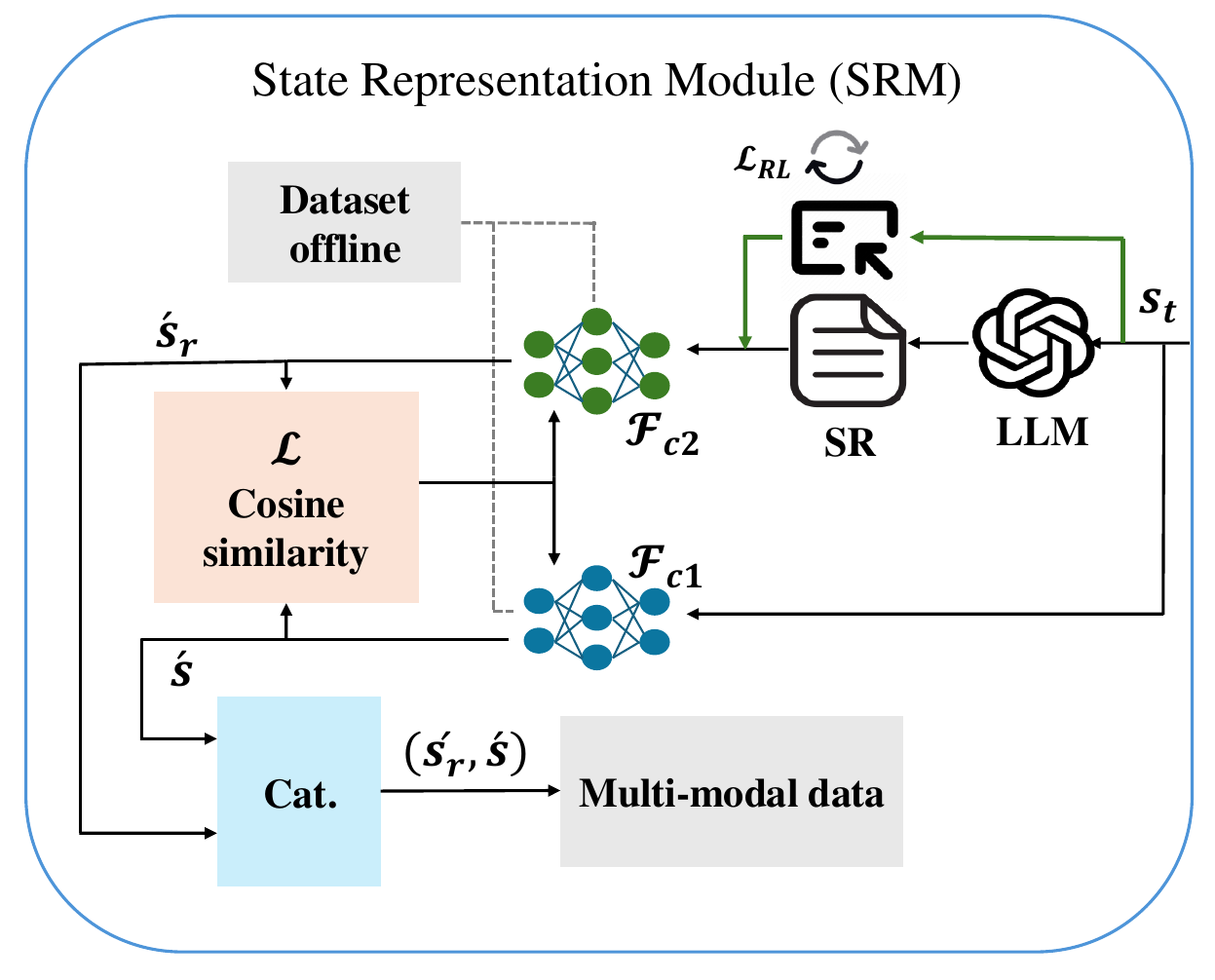}
  \caption{SRM structure with prompt learner integration.
  }\vspace{-0.2cm} 
  \label{fig:srm}
  \end{figure}
Efficient dynamic resource allocation is critical in O-RAN due to the highly dynamic wireless environment and time-varying network conditions. The RIC leverages ML approaches, improving adaptability in resource management.
For details on the OFDM-based transmission scheme and UE-level data rate formulation, we refer the reader to our previous work\cite{lotfi2025llm}. 

\vspace{-0.2cm}
\subsection{QoS Metrics for Slice Performance Evaluation}
The QoS performance for each slice is defined by specific KPIs tailored to their objectives:

\noindent\textbf{Slice 1 }(eMBB - High Throughput Focus):
The primary KPI is aggregate user throughput, ensuring high data rates across UEs as 
$\mu_r = \frac{1}{N_u}\sum_{i=1}^{N_u} C_i,$
where $C_i$ denotes the throughput for user $i$, and $N_u$ is the total number of UEs.

\noindent\textbf{Slice 2 }(mMTC - Device Density and Availability):
The KPI combines user throughput and UE density, ensuring reliable connectivity for a massive number of devices as 
$\textit{d}s = \frac{\sum_{i=1}^{N_u}\mathds{1}{(C_i > \lambda_i)}}{N_u}\sum_{i=1}^{N_u} C_i,$
where $\mathds{1}$ is an indicator function equal to 1 if $C_i(t) \geq \lambda_i$ and 0 otherwise.

\noindent\textbf{Slice 3 }(URLLC - Latency Minimization):
Latency-critical applications require minimizing the worst-case transmission delay as 
$\textit{l}_d = \max (\tau_i), \forall i \in N_u,$
where $\tau_i$ represents the latency for UE $i$.

\noindent The overall QoS vector for all slices is defined as
$Q^l = \boldsymbol{Q}[l]$, where $\boldsymbol{Q} = [\mu_r, d_s, l_d], \forall l \in \{1, 2, 3\}$. \vspace{-0.2cm}
\subsection{Problem Formulation: Maximizing Multi-Slice Performance}

To achieve an optimal balance between performance and fairness across slices, we adopt a weighted utility-based optimization strategy, where the utility function $U^l$ represents the QoS performance of slice $l$. 
This optimization is formulated as follows:\vspace{-0.2cm}
\begin{subequations}
\begin{align}\label{opt1}
\argmax_{\boldsymbol{b}, \boldsymbol{e}} & \hspace{0.5cm}
\sum_{l \in \mathcal{\mathcal{L}}} w_l U^l(\boldsymbol{b}, \boldsymbol{e}) \\
\text{s.t.,}
& \hspace{0.5cm} \boldsymbol{b}_l \in {0,1}^{|\mathcal{L}| \times K^l_m}, \quad \boldsymbol{e}_l \in {0,1}^{N{u,m} \times K^l_m}, \\
& \hspace{0.5cm} \sum_{l=1}^{|\mathcal{L}|} \sum_{u=1}^{N_{u,m}} \sum_{k=1}^{K_m} e_{u,k} b_{l,k} \leq K_m, \label{opt_new_rb}\\
& \hspace{0.5cm} \sum_{l} b_{l,k} \leq 1 + \lambda_l \max(0, \sum_l b_{l,k} - 1), \label{opt_new_e}\\
& \hspace{0.5cm} Q^l_{\boldsymbol{b}_l, \boldsymbol{e}_l} \geq Q^l{min} - \delta_l, \label{opt_new_qos}
\end{align}
\end{subequations}
where $w_l$ determines the priority weight of slice $l$ in the optimization objective and $N_{u,m}$ denotes the number of user equipments (UEs) served by a given RU-DU pair. 
The constraint in \eqref{opt_new_rb} ensures that the total assigned RBs do not exceed the available capacity $K_m$ at a given RU-DU pair $m$. \eqref{opt_new_e} introduces a relaxed allocation condition with a penalty term $\lambda_k$ to enable soft sharing and exploration, while final allocations enforce one-slice-per-RB assignments. 
Finally, \eqref{opt_new_qos} ensures that each slice maintains a minimum QoS requirement $Q^l_{min}$, with $\delta_l$ allowing controlled flexibility to prevent infeasibility in resource allocation. The combination of a weighted utility function, relaxed RB allocation constraints, and adaptive QoS guarantees enables a more flexible and scalable approach to dynamic O-RAN management.

To address this, we reformulate the problem as a Markov Decision Process (MDP), enabling RL to dynamically optimize resource allocation. In this MDP framework, the state space captures network conditions, including resource availability and QoS requirements, while the action space represents allocation decisions. 
To enhance optimization, we leverage DRL techniques to learn optimal allocation strategies. 
To address these challenges, we introduce a prompt-augmented learning approach that complements DRL by integrating structured guidance into the decision-making process. This enables the system to achieve more robust and adaptive slicing policies while ensuring stable and optimized resource management. 
The following section details our proposed approach, outlining the DRL formulation and the integration of LLM-driven prompt learning in optimizing network slicing decisions. \vspace{-0.2cm}

\section{Promp-Augmented Multi Agent RL (PA-MRL)}\label{padrl}

\subsection{MDP Problem}
To tackle the optimization challenge in \eqref{opt1}, we formulate it as a MDP problem, represented by the tuple $\langle \mathcal{S}, \mathcal{A}, T, \gamma, r \rangle$. Here, $T$ denotes the 
transition probability from state $s_t$ to $s_{t+1}$ given action $a_t$, and $\gamma$ represents a discount factor in $[0,1]$ to balance short-term and long-term rewards. The remaining components of the MDP are outlined below. 

\textbf{(1) State Space ($\mathcal{S}$):}
The state representation at time step $t$ captures the current network status, providing the agent with essential information for decision-making. Specifically, the state $s_t \in \mathcal{S}$ includes the QoS levels of each slice, user density of each slice, and the previous resource allocation decision, formally expressed as $s_t = \{ Q^l, N^l_u, a_{t-1} \mid \forall l \in \mathcal{L} \}$.

\textbf{(2) Action Space ($\mathcal{A}$):}
At each decision step, the agent selects an action $a_t \in \mathcal{A}$, which determines the resource allocation strategy across slices. The action space consists of two primary allocation variables: inter-slice resource allocation ($\boldsymbol{b}$) and intra-slice resource assignment ($\boldsymbol{e}$). Thus, the action taken at time $t$ is given by $a_t = \{ \boldsymbol{b}, \boldsymbol{e} \}$.

\textbf{(3) Reward Function ($r$):}
The reward function $r_t$ evaluates the effectiveness and fairness of resource allocation across slices. Each slice $l$ is associated with a QoS metric $Q^l$ (e.g., throughput, latency, or connection density) and its corresponding target threshold $\text{thr}^l$. To ensure consistency across heterogeneous metrics, we define reward contributions based on the normalized deviation of $Q^l$ from $\text{thr}^l$:\vspace{-0.1cm}
\begin{equation}
r_0^l = \frac{1}{1 + \exp\left(-\alpha \cdot \left(\frac{Q^l - \text{thr}^l}{\text{thr}^l}\right)\right)},
\label{eq:r0}
\end{equation}
where $\alpha$ controls the steepness of the reward transition around the threshold. This formulation provides a smooth and differentiable signal that incentivizes the agent to meet or exceed the target QoS for each slice. The total reward from all slices is aggregated as $r_Q = \sum_{l \in \mathcal{S}} r_0^l$, and an exponential penalty is applied for severe violations:
\begin{equation}
\text{penalty} = \sum_{l: Q^l < \text{thr}^l(1 - \text{margin})} \exp\left(-\delta \cdot \frac{Q^l - \text{thr}^l}{\text{thr}^l}\right),
\label{eq:penalty}
\end{equation}
where $\delta$ adjusts the severity of the penalty. The final reward is then given by $r_t = r_Q - \text{penalty}$. 
Building on this formulation, the goal of the agent is to maximize the cumulative expected reward over time, $R(t) = \sum_{i=0}^{\infty} \gamma^i r_{i,t}$. To achieve this, the agent learns an optimal policy $\pi^*(a_t \mid s_t; \theta_p)$, parameterized by $\theta_p$, which guides the action-selection process based on the observed state $s_t$.\vspace{-0.1cm} 


\subsection{Integration of LLM-Driven Decision Support}
To effectively solve the formulated MDP and improve decision-making under uncertainty, we incorporate LLMs into our DRL framework. Traditional DRL approaches struggle with stability and generalization, particularly in high-dimensional state spaces like O-RAN slicing. By leveraging ORANSight LLM~\cite{gajjar2025oransight} model that have fine-tuned on extensive O-RAN knowledge and prompt-augmented learning, our framework integrates two complementary prompting mechanisms:
\begin{enumerate}[left=0pt]
\item \textbf{Informal LLM Prompts}, which introduce scenario-aware guidance to enhance decision flexibility, allowing the DRL agent to adapt more effectively to complex and dynamic network environments.
\item \textbf{Learnable Prompts}, which optimize policy representations by embedding structured knowledge into the DRL framework, improving generalization and stability across diverse slicing scenarios.
\end{enumerate}
By embedding these prompt-driven adaptations within the learning pipeline, our approach enhances robustness, accelerates convergence, and improves real-time network slicing decisions. The high-level interaction between the RL agent (actor/critic), SRM, and near-RT RIC is shown in Fig.~\ref{sys_graph}, while the internal structure of the SRM is detailed in Fig.~\ref{fig:srm}. 

\subsection{Proposed Prompt-Augmented MARL approach }

Building upon this LLM integration, this work proposes a contextualization-based adaptation method that enhances DRL for dynamic O-RAN network slicing by integrating learnable prompts into a pre-trained LLM. Rather than fully fine-tuning the LLM, which is computationally expensive and often leads to overfitting, our approach introduces task-specific prompts that dynamically guide the model in real-time decision making. These prompts enable the LLM to interpret complex network states, improve the agent’s policy adaptation, and facilitate efficient resource allocation without altering the core parameters of the pretrained model.
Our framework leverages two complementary types of prompts to bridge the gap between numerical network data and semantic understanding:

\textbf{Informal Prompts} provide high-level contextual guidance by converting network parameters, such as RF conditions, QoS metrics, and traffic loads, into structured natural language descriptions. These prompts enable the LLM to capture and interpret relationships within the environment that might be overlooked in purely numerical representations. A typical prompt might be structured as: “Slice 1 has a QoS level of \textless\ \textgreater\ and a throughput of \textless\ \textgreater\ Mbps,” providing a semantic representation of network states.

\textbf{Learnable Prompts} enhance semantic extraction and policy refinement by introducing a small set of trainable tokens that interact with the LLM and the DRL agent. These tokens adapt dynamically based on the agent’s observations, refining decision-making by aligning the model’s representation with slice-specific optimization needs. 

This design is particularly beneficial in dynamic environments, where unstructured feedback often challenges the ability of DRL agents to form consistent policies. 
To address these challenges, we introduce a contextualization-based adaptation method that integrates learnable prompts into LLM-augmented DRL, enhancing state representation and accelerating learning. In a typical RL framework, the agent interacts with the environment, receiving scattered and dynamic feedback that takes time to become meaningful. LLMs play a crucial role by mapping these scattered prompts into structured latent representations, effectively clustering related concepts in their latent space. For example, in O-RAN slicing, the idea of SNR is naturally related to power levels, and the LLM's internal structure captures these relationships, providing a more cohesive and interpretable state representation. Instead of treating feedback as isolated data points, the LLM clusters semantically similar information, making it easier for the RL agent to extract patterns and optimize resource allocation.
We develop ORANSight, a specialized LLM pre-trained in O-RAN knowledge, and integrate it into a Prompt-Augmented Multi-Agent RL (PA-MRL) framework. Within this framework, learnable prompts are added to the state representation, allowing the model to dynamically refine how it interprets network conditions. Unlike static LLM embeddings, these prompts are optimized through training to maximize RL rewards, ensuring that the state representations evolve to improve decision-making. This dual training constraint, aligning representations to both meaningful clustering and RL objectives, enables the RL agent to achieve higher rewards with fewer training iterations, improving efficiency and adaptability in O-RAN slicing.
By incorporating prompt-augmented learning into MARL, our approach enhances the ability of the RIC to make more informed, scalable, and adaptive resource allocation decisions, ultimately improving the efficiency and flexibility of O-RAN slicing. Experimental results demonstrate that our method accelerates convergence, reduces computational overhead, and outperforms traditional fine-tuning in efficiency, scalability, and adaptability, making it an ideal solution for real-world O-RAN deployment.

By combining informal textual prompts with learnable token adaptation, our approach preserves the pretrained knowledge of the LLM while significantly improving its interpretability and adaptability for O-RAN environments. This prompt-augmented learning strategy enables the DRL agent to leverage contextual insights, leading to more efficient, robust, and generalizable network-slicing policies. 
Building on our previous work~\cite{lotfi2025llm}, where informal prompts were utilized to transform network parameters into structured textual representations for LLM-based decision support, we extend this approach by introducing learnable prompts to enhance the model's adaptability and semantic extraction capabilities. Instead of relying solely on static textual descriptions, we incorporate a set of trainable tokens that dynamically interact with the LLM, refining its understanding of network conditions in real-time. 
To align these modalities into a unified feature space, we employ two pre-trained adapter networks, $\mathcal{F}{c_1}$ and $\mathcal{F}{c_2}$, trained offline using real state data $s_t$ and corresponding LLM-generated representations $s_{r,t}$. These adapter networks are based on our previous work~\cite{lotfi2025llm}, enabling cohesive processing of textual and numerical state data.

\begin{algorithm}[t!]
\SetAlgoLined
\textbf{Input}: Number of iterations $N_t$, actors $N_m$, evaluations $N_e$, actor weights $\theta_{p,i}$, critic weights $\theta_{v}$, pre-trained LLM $\mathcal{M}$, informal prompt set $\mathcal{P}$, learnable prompt embeddings $\mathcal{T}$, adapter networks $\mathcal{F}_{c1}, \mathcal{F}_{c2}$. \\
\textbf{Initialize:} Initialize $\theta_{p,i}$, $\theta_v$, and $\mathcal{T}$. \\
\For{iteration $t=1:N_t$}{
    \For{actor $i=1:N_m$}{
        Construct prompt $p_t \in \mathcal{P}$ from state $s_t$ via network metrics.\\
        Feed $p_t \cup \mathcal{T}$ to LLM.\\
        $h_t \gets \mathcal{M}(p_t \cup \mathcal{T})$ \\ 
        $s'_{r,t} \gets \mathcal{F}_{c2}(h_t)$, $s'_t \gets \mathcal{F}_{c1}(s_t)$ \\
        $r_i = \text{evaluate}(\pi_{p,i}(s'_{r,t}, s'_t))$ \\
        $\mathcal{B} \gets \langle (s'_{r,t}, s'_t), a_t, (s'_{r,t+1}, s'_{t+1}), r_t \rangle$ \\
    }
}
Update global critic network $\theta_v$ using \eqref{vupdate}. \\
Update $N_m$ actor networks $\theta_{p,i}$ using \eqref{pupdate}. \\
Update $\mathcal{T}$ via RL gradients.\\
\If{$\theta_{p, \forall i}$ converges}{
    Break.
}
\textbf{Output}: Trained policy weights $\theta_{p,i}$, updated critic weights $\theta_{v}$, optimized prompt embeddings $\mathcal{T}$ \\
\caption{Prompt-Augmented MARL (PA-MRL) Algorithm}
\label{alg1}
\end{algorithm}

In the O-RAN slicing scenario, the RIC employs the PA-MRL framework, integrating both learnable and informal prompts to enhance adaptability in network slicing and resource management. By leveraging this hybrid prompting strategy, PA-MRL enables the system to dynamically interpret network conditions, optimize resource allocation, and adjust to varying traffic demands in real time. 
This approach offers several key advantages: more generalizable state representations, improved adaptation to diverse network slicing scenarios, and an end-to-end differentiable pipeline. Rather than relying on rigid, predefined templates, the learnable tokens evolve throughout training, creating flexible, context-aware representations of network states. Additionally, learnable token embeddings can be optimized for different slicing configurations, allowing the model to refine its understanding and decision-making based on specific deployment scenarios. Finally, by integrating learnable prompts within the DRL framework, the system continuously improves how it processes textual representations through real-time training feedback, making it more responsive and adaptive to changing network conditions. \vspace{-0.2cm}





\subsection{Soft Actor-Critic in a Multi-Agent Setup}

To implement effective policy learning within our proposed PA-MRL framework, we adopt the Soft Actor-Critic (SAC) algorithm within a multi-agent RL (MARL) framework. SAC is well-suited for environments with continuous action spaces and dynamic feedback, as it jointly maximizes expected return and policy entropy, encouraging both stability and exploration~\cite{haarnoja2018soft}. 
Our architecture deploys multiple distributed SAC agents across O-DUs to improve scalability and sample efficiency. Each agent locally interacts with the environment, while a centralized critic located in the O-CU within the near-RT RIC, aggregates experiences and updates policy and value networks using gradient-based learning.

The policy network update follows the entropy-regularized objective:\vspace{-0.1cm}
\begin{align}
\nabla_{\theta_p}J(\pi_{\theta_p}) &= \mathbb{E}_{\kappa,\pi} \Big[ \nabla_{\theta_p} \log \pi_{\theta_p}(a_t | s_t) \nonumber \\
&\big( -\beta \log \pi_{\theta_p}(a_t | s_t) + Q_v(s_t, a_t) \big) \Big],
\label{pupdate}
\end{align}
where $\beta$ controls the balance between reward maximization and entropy regularization. The Q-value function is defined as $Q_v(s_t, a_t) = \mathbb{E}_{\pi}[R(t) \mid s_t, a_t]$. 
The value network parameters $\theta_v$ are updated by minimizing the temporal-difference loss:
\begin{align}
\min_{\theta_v} \mathbb{E}_{\kappa,\pi} \left[\left(y_t - Q_v(s_t, a_t; \theta_v)\right)^2 \right],
\label{vupdate}
\end{align}
with the target value computed as $ y_t = r_t + \gamma Q_v(s_{t+1}, a_{t+1}; \theta_v) - \beta \log \pi_{\theta_p}(a_{t+1} | s_{t+1})$. \vspace{-0.1cm}



\section{Evaluation Results }\label{simulation}
\subsection{Simulator and Parameter Settings}
We simulate an O-RAN environment with three slices, eMBB, MTC, and URLLC, distributed across $N_m = 6$ DUs, each acting as an independent agent managing varying user traffic and service demands. Users ($N_u = {50, 100, 200}$) are uniformly distributed and move at $10$–$20$ m/s in one of seven possible directions $\{\pm \pi/3, \pm \pi/6, \pm \pi/12, 0\}$, following the mobility pattern in~\cite{lotfi2025llm}. Key parameters include $15$ kHz subcarrier spacing, $20$ MHz bandwidth per DU, $200$ kHz RB size, and $p_u = 56$ dBm user transmission power~\cite{3gpp15}, under Rayleigh fading and additive white noise.
We implement our PA-MRL strategy using an actor-critic architecture in PyTorch, with three fully connected layers ($600$, $700$, and $700$ neurons) and \textit{tanh} activation. All models are trained with the \textit{Adam} optimizer at a learning rate $1e-4$ and a batch size of $128$. 
We evaluate three scenarios: (1) ORANSight PA-MRL (proposed), which integrates a domain-specific LLM with prompt-aligned SRMs to provide context-aware insights at both the actor and critic levels; (2) GPT PA-MRL, which replaces ORANSight with GPT-2 while retaining prompt alignment; and (3) ORANSight MARL, which uses the ORANSight LLM within the SRM but omits prompt alignment~\cite{lotfi2025llm}. All scenarios apply SRMs at both the DU and critic levels, and the prompt generation process builds upon framework in~\cite{lotfi2025llm}. \vspace{-0.2cm}

\subsection{Convergence Performance}

\begin{figure}[t!]
  \centering
\includegraphics[width=0.65\columnwidth]{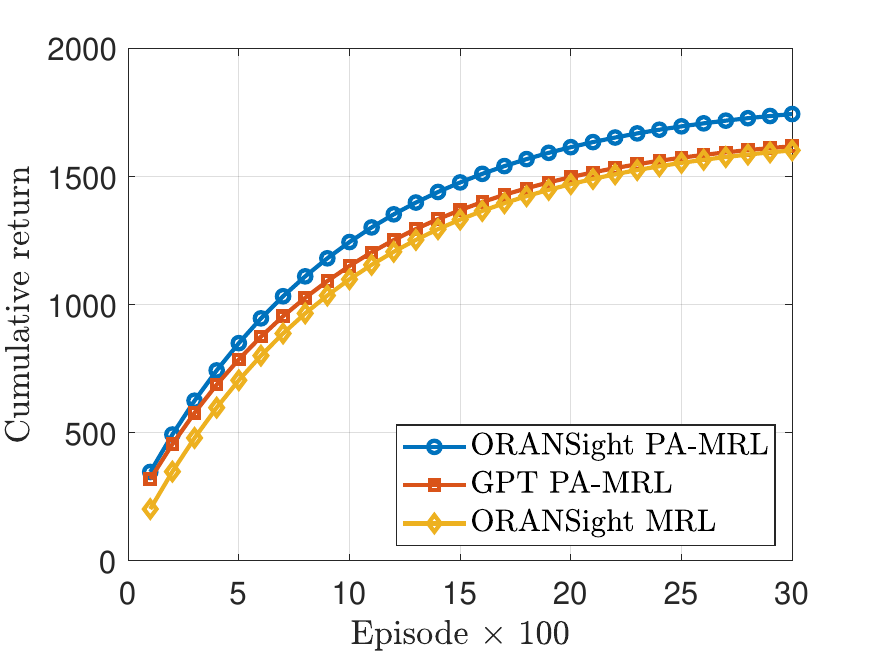}\vspace{-0.2cm}
    \caption{\small Average cumulative reward values achieved by various approaches. 
    }\vspace{-0.2cm}
    \label{creward}
\end{figure}

Fig.~\ref{creward} presents the cumulative reward across training episodes for three methods: the proposed ORANSight PA-MRL, GPT PA-MRL, and the baseline ORANSight MRL. As shown, ORANSight PA-MRL consistently outperforms the other two approaches, achieving faster convergence and a higher final cumulative reward. The performance gap becomes more pronounced after 1000 episodes, demonstrating the advantage of domain-specific LLM model and integrating task-based prompt alignment in LLM-based state representation modules within the actor and critic. GPT PA-MRL, which utilizes the GPT-2 model in the SRM module, shows moderate improvement over the baseline but lags behind ORANSight PA-MRL, highlighting the benefit of prompt alignment in policy optimization. \vspace{-0.2cm}%


\subsection{Number of Context} 
\begin{figure}[t!]
  \centering
\includegraphics[width=0.65\columnwidth]{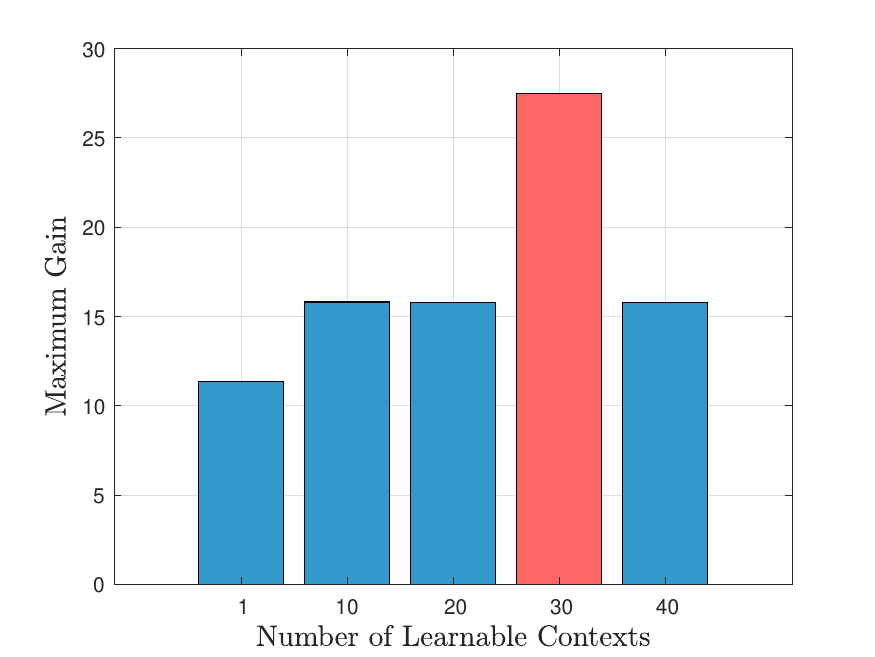}\vspace{-0.2cm}
    \caption{\small Comparison of maximum reward gains under varying numbers of learnable context.
    }\vspace{-0.2cm}
    \label{cntnumb}
\end{figure}

The performance analysis based on the number of context tokens (or added tokens) is shown in Fig.~\ref{cntnumb}. A large number of tokens may lead to overfitting, while a small number can result in underfitting. Therefore, this value should be treated as a tunable hyperparameter depending on the application. The dominant maximum is highlighted in red to indicate the peak performance. 

\begin{table}[t!] 
\footnotesize
\centering
\caption{Ablation Study Results}\vspace{-0.2cm}
\resizebox{\columnwidth}{!}{%
\begin{tabular}{|>{\centering\arraybackslash}m{2.4cm}|>{\centering\arraybackslash}m{1.2cm}|>{\centering\arraybackslash}m{1.2cm}|>
{\centering\arraybackslash}m{1.2cm}|>{\centering\arraybackslash}m{1.7cm}|}
\hline
\textbf{Model Version}              & \textbf{QoS (eMBB)} & \textbf{QoS (mMTC)} & \textbf{QoS (URLLC)} & \textbf{Convergence} \\ \hline
\textbf{ORANSight PA-MRL}  & 55.4\%                & 47.05\%                 & 23.66\%                   & 8.97\%                \\ \hline
\textbf{GPT PA-MRL}                & 11.14\%                & 35.2\%                 & 17.15\%                   & 1.69\%              \\ \hline
\textbf{ORANSight MARL}                & 2.6\%                & 23.5\%                 & 7.27\%                   & 1.12\%              \\ \hline
\end{tabular}%
}\label{table:ablation} \vspace{-0.3cm}
\end{table}
\subsection{Ablation Study}
Table~\ref{table:ablation} presents the results of an ablation study comparing the proposed ORANSight PA-MRL framework with its reduced variants: GPT PA-MRL, which replaces the domain-specific LLM with a general-purpose GPT-2 model in the state representation module; and ORANSight MARL, which removes the prompt-aligned LLM integration entirely. These are evaluated against a plain MARL baseline.  
The proposed ORANSight PA-MRL achieves the best overall performance, confirming the impact of domain-specific LLMs and prompt alignment in both actor and critic. GPT PA-MRL shows moderate improvement, highlighting the role of LLM integration and the limitations of generic models. The baseline ORANSight MARL lags behind, validating the benefit of prompt-based state representation in learning stability and slice-level QoS. 
\vspace{-0.1cm}

\subsection{Network users' QoE}

\begin{figure}[t!]
  \centering
    \includegraphics[width=0.65\columnwidth]{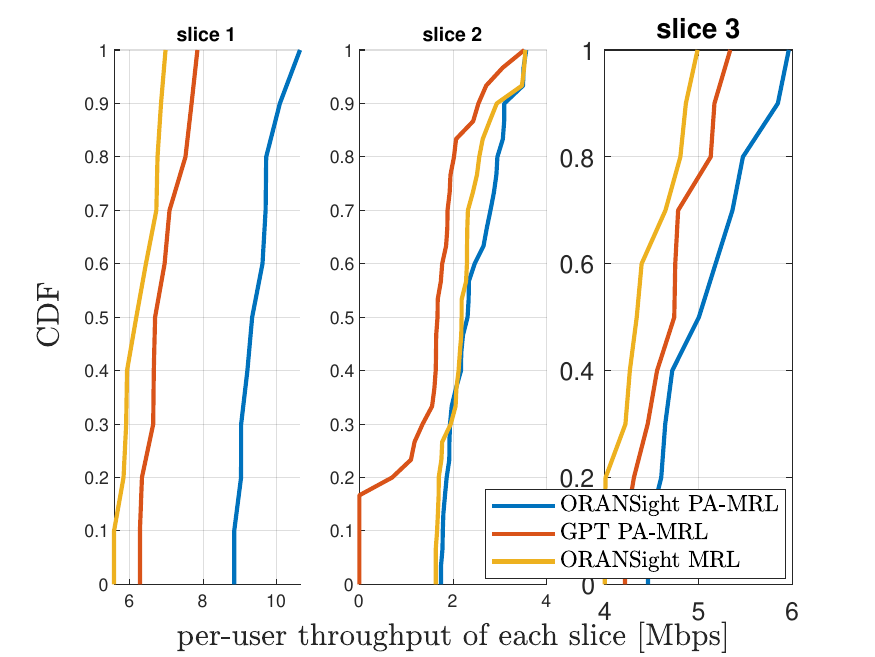}\vspace{-0.1cm}
    \caption{\small CDF of UEs throughput across the network. }\vspace{-0.2cm}
    \label{uethr}
\end{figure}

Fig.~\ref{uethr} shows the per-user throughput distribution across slices. ORANSight PA-MRL consistently shifts the CDF curves to the right, indicating higher throughput and improved fairness across UEs. GPT PA-MRL offers a modest improvement over the baseline, while ORANSight MRL results in lower per-user rates, especially in slice 2. These results highlight the effectiveness of prompt-aligned LLM-based state representations in achieving both efficiency and user-level QoS guarantees.

\section{Conclusion}\label{conclusion}
The dynamic nature of next-generation wireless networks makes resource management and network slicing challenging. To address this issue, this paper proposed a novel task-related prompt-augmented LLM-based multi-agent reinforcement learning (PA-MRL) framework for O-RAN network slicing. By incorporating a prompt-alignment state representation module into actor and critic networks and utilizing a domain-specific LLM pre-trained model, the proposed ORANSight PA-MRL enables more expressive and task-aware policy learning. Experimental results demonstrate significant improvements in cumulative reward, slice-level QoS, and convergence performance in both general-purpose LLM models and non-LLM baselines. 
The proposed framework presents a promising step toward scalable, intelligent resource management in next-generation wireless networks. 
\vspace{-0.1cm}

\def\baselinestretch{0.81}
\bibliographystyle{IEEEbib}
\bibliography{Main}
\end{document}